\documentclass{article}
\usepackage{graphicx} 
\usepackage{CJKutf8}
\usepackage{amsmath}
\usepackage{booktabs} 
\usepackage{natbib} 
\usepackage{authblk}
\usepackage{amssymb}
\usepackage{hyperref}
\usepackage[final]{neurips}
\usepackage{cite}

\begin{document}

\title{Scaling Towards the Information Boundary of Instruction Sets: The Infinity Instruct Subject Technical Report}
\author{LiDu, Hanyu Zhao, Yiming Ju, Chengwei Wu, Tengfei Pan }



\author{
  Li Du$^{1,2}$, \ 
  Hanyu Zhao$^{1,2}$, \
  Yiming Ju$^{1,2}$, \
  Chengwei Wu$^{1}$, \
  Tengfei Pan \\
  \text{BAAI} \\ \\
  {\{ldu, hyzhao, ymju, cwwu, tfpan\}@baai.ac.cn
}
}


\date{\today}

\maketitle

\begin{CJK}{UTF8}{gbsn}


\section{Abstract}

Instruction tuning has become a foundation for unlocking the capabilities of large-scale pretrained models and improving their performance on complex tasks. Thus, the construction of high-quality instruction datasets is crucial for enhancing model performance and generalizability. Although current instruction datasets have reached tens of millions of samples, models finetuned on them may still struggle with complex instruction following and tasks in rare domains. This is primarily due to limited expansion in both ``coverage'' (coverage of task types and knowledge areas) and ``depth'' (instruction complexity) of the instruction set.
To address this issue, we propose a systematic instruction data construction framework, which integrates a hierarchical tagging system, an informative seed selection algorithm, an evolutionary data synthesis process, and a model deficiency diagnosis with targeted data generation. These components form an iterative closed-loop to continuously enhance the coverage and depth of instruction data.
Based on this framework, we construct Infinity Instruct Subject, a high-quality dataset containing $\sim$1.5 million instructions. Experiments on multiple foundation models and benchmark tasks demonstrate its effectiveness in improving instruction-following capabilities. Further analyses suggest that Infinity Instruct Subject shows enlarged coverage and depth compared to comparable synthesized instruction datasets. Our work lays a theoretical and practical foundation for the efficient, continuous evolution of instruction datasets, moving from data quantity expansion to qualitative improvement. 

\footnotetext[1]{These authors contributed equally.}
\footnotetext[2]{Corresponding author.}

\section{Introduction}


Instruction tuning serves as a cornerstone for unleashing the vast knowledge and reasoning capabilities of large pretrained models \citep{longpre2023flan,ouyang2022training}. Consequently, the construction of high-quality instruction datasets has become essential for improving model performance and generalization ability. Several efforts have constructed instruction datasets through manual annotation or automatic synthesis, with the total size of these datasets having reached tens of millions \citep{ahmad2025opencodeinstruct, nayak2024learning}. However, current large models still struggle with complex instruction-following tasks and show limitation in ``rare'' tasks with low frequency \citep{qin2025incentivizing}. 


A main reason would be that existing works overlooked the distribution of existing data, and thus failed to efficiently enhance the ``coverage and ``depth of synthesized instructions, and thus limit the model performance in solving rare tasks and difficult tasks \citep{wang2024survey,liu2024best,long2024llms}.
The ``coverage of an instruction dataset refers to the coverage of task types and knowledge categories it covers, while the `depth reflects the complexity of instructions within individual tasks. Both two dimensions are essential to enlarge the capability boundaries of large models \citep{shengyu2023instruction}.
Expanding the coverage improves the model’s ability to generalize across domains, while increasing depth helps enhance complex reasoning ability and compositional generalizability \citep{sinha2024survey,zhong2025survey}. However, a key challenge remains: how to design \textbf{effective, controllable and interpretable} instruction generation strategies that continuously expand the instruction space along both dimensions, guiding models to acquire higher-level capabilities \citep{zhang2023survey,qian2022controllable,qin2022cold}. Currently, there is a lack of unified methodology and empirical research in this area.

\begin{figure*}[ht]
    \centering
    \includegraphics[width=0.9\linewidth]{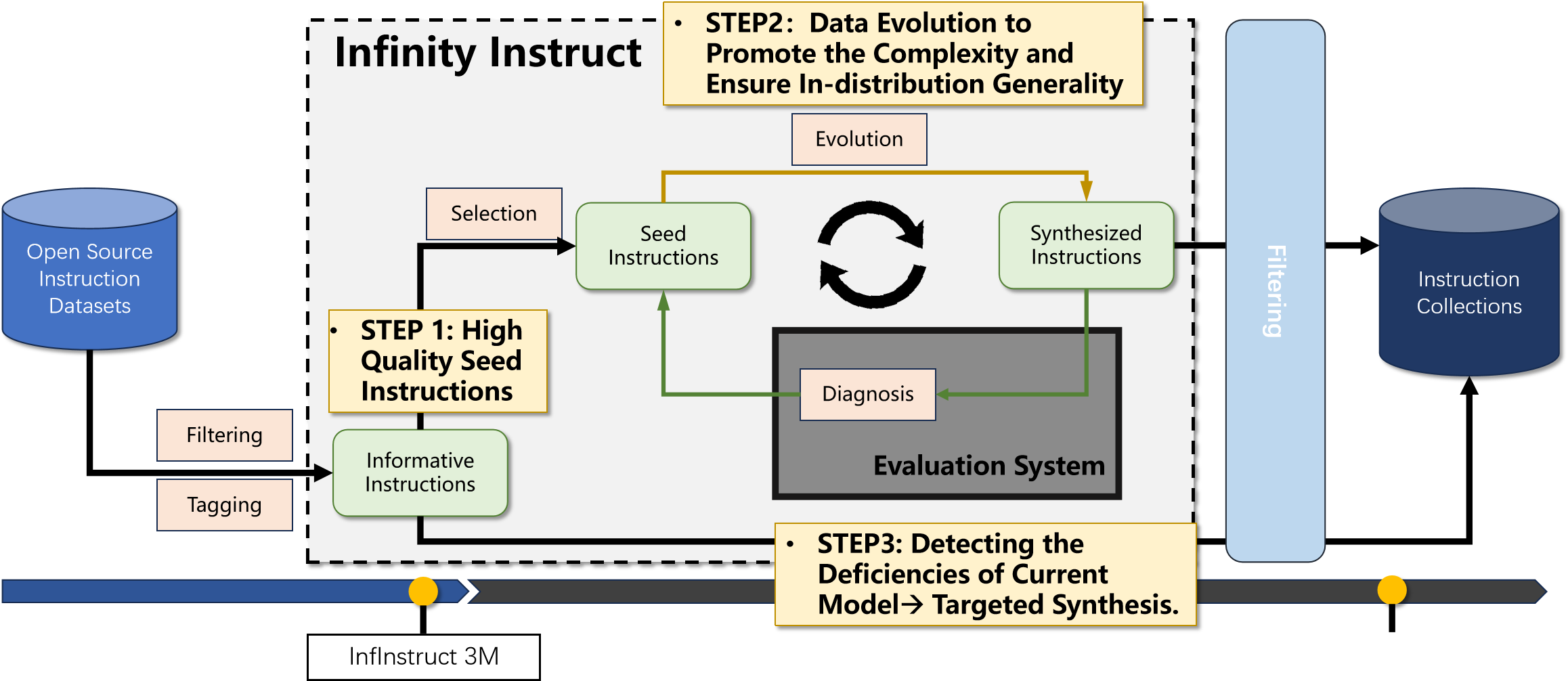}
    \caption{Construction pipeline of the InfInstruct-Sub dataset.  }
    \label{fig:frame}
\end{figure*}

To address this challenge, we propose a comprehensive instruction data construction framework, which goal is to measure the current data distribution and synthesize instructions that target under-covered and high-complexity regions, thereby expanding both the coverage and depth of the dataset. This framework consists of four core components:
(1) a hierarchical and multilingual tagging system to understand the content and ability distribution of existing instruction content;
(2) informative seed data selection to identify valuable instructions with low coverage or high difficulty;
(3) evolutionary data synthesis to synthesize more complex instructions by evolving from seed data;
(4) a model deficiency diagnosis based targeted synthesis module to identify potential flaws in the model’s knowledge or capabilities and synthesize data to address those weaknesses.
As illustrated in Figure~\ref{fig:frame}, these four modules form a closed-loop system that can iteratively expand the coverage and depth of the instruction dataset.


Based on this framework, we construct the Infinity Instruct Subject (InfInstruct-Sub) dataset, which contains $\sim$1.5 million high-quality instructions. The code and dataset are available at \url{https://github.com/BAAI-DIPL/InfinityInstruct-Sub} and \url{https://huggingface.co/datasets/BAAI/Infinity-Instruct/tree/main/Gen}. Evaluations on multiple benchmarks demonstrate its effectiveness in improving instruction-following ability. Across foundation models, our tuned models outperform their official instruction-tuned counterparts. Data analysis further reveals that, compared with comparable synthetic datasets such as Magpie \citep{magicoder} and UltraChat \citep{ding2023enhancing}, our framework could obtain an instruction set with higher coverage and complexity. Moreover, interestingly, we observe a scaling law in the co-occurrence structure of tags of instructions: the co-occurrence frequency of a tag with others follows a negative logarithmic relationship with its own frequency. This indicates a scale-free connection structure between instructions, similar to the topology of the Internet. Such a pattern may offer new insights into the scaling laws of model performance and suggest ways to better understand the internal knowledge structure of datasets, potentially improving training efficiency and effectiveness. This further highlights the importance of understanding the content structure of instruction data, and underscores the value of the high-quality tagging system proposed in this work.



\section{Construction of Infinity Instruct Subject}


The construction process of Infinity Instruct Subject (InfInstruct-Sub) is illustrated in Figure~\ref{fig:frame}. The process starts by collecting high-quality seed instructions. We systematically gathered all general-domain instruction datasets both manually created and automatically generated using GPT-4\citep{achiam2023gpt} or ChatGPT available up to March 2024, resulting in a data pool of approximately 7 million samples. The included datasets are listed in Table~\ref{tab:dataset_summary} of the Appendix. 
Next, we developed an automatic tagging system based on large language models (LLMs) to analyze the distribution of the data pool. From the perspectives of coverage and depth, we selected a set of high-information seed instructions. 
Then, we applied an evolutionary algorithm to generate over one million new instruction samples by evolving towards greater complexity and difficulty.
Building on both the seed and synthesized data, we constructed a model deficiency diagnosis system. This system identifies gaps in model capabilities and guides the targeted synthesis of new data to efficiently address those weaknesses.
Finally, we implemented a strict semantic similarity–based data leakage prevention framework to detect and mitigate potential risks of data leakage throughout the construction process.




\subsection{Hierarchical Multilingual Tagging System for Characterizing Instruction Content Distribution}

\begin{figure*}
    \centering
    \includegraphics[width=0.95\linewidth]{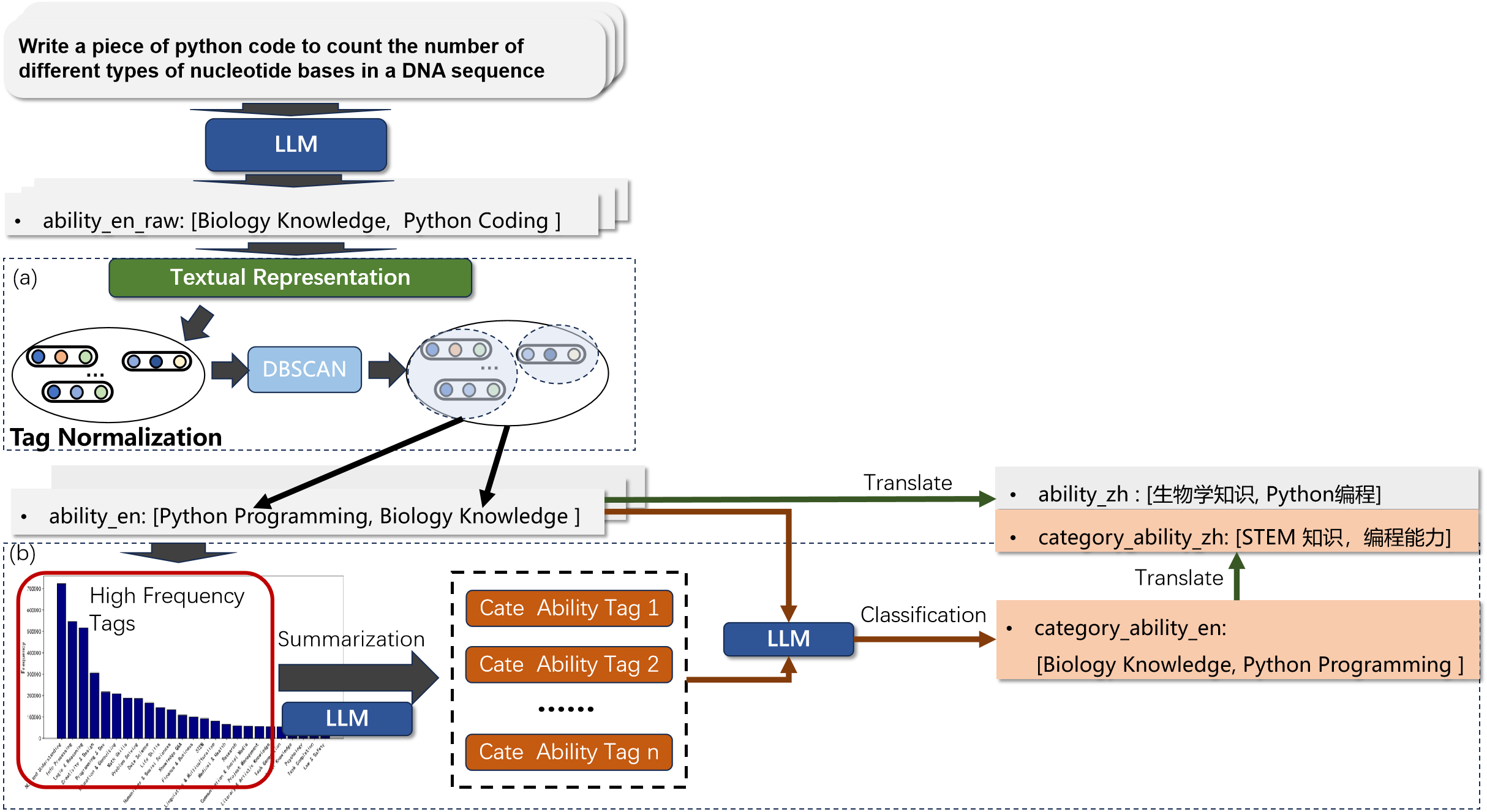}
    \caption{Tagging system of the InfInstruct-Sub dataset for elucidating content distribution of instruction pools. (a) Fine-grained tags and normalization of fine-grained tags. (b) Construction of categorical tags and the process of mapping fine-grained tags to categorical tags.}
    \label{fig:tag_system}
\end{figure*}

A prerequisite for synthesizing instructions that target under-covered areas or exhibit high complexity is understanding the content distribution of existing instruction data. To this end, we design a hierarchical, multilingual tagging system powered by LLMs as illustrated in Figure~\ref{fig:tag_system}. This system assigns each instruction both Chinese and English tags at two levels: domain-level tags and fine-grained tags. These tags indicate the domain of each instruction, as well as the types of abilities and knowledge required to complete it, facilitating interpretation and selection for downstream tasks.

Specifically, the tagging process is implemented in a bottom-up manner. We first generate fine-grained tags for each instruction. These are then normalized (Figure~\ref{fig:tag_system}(a)) to remove noise and merge semantically equivalent tags expressed in different forms. Finally, LLMs are used to automatically cluster and abstract fine-grained tags into broader domain-level categories, and to map each fine-grained tag to its corresponding domain tag (Figure~\ref{fig:tag_system}(b)).

\begin{figure*}
    \centering
    \includegraphics[width=0.8\linewidth]{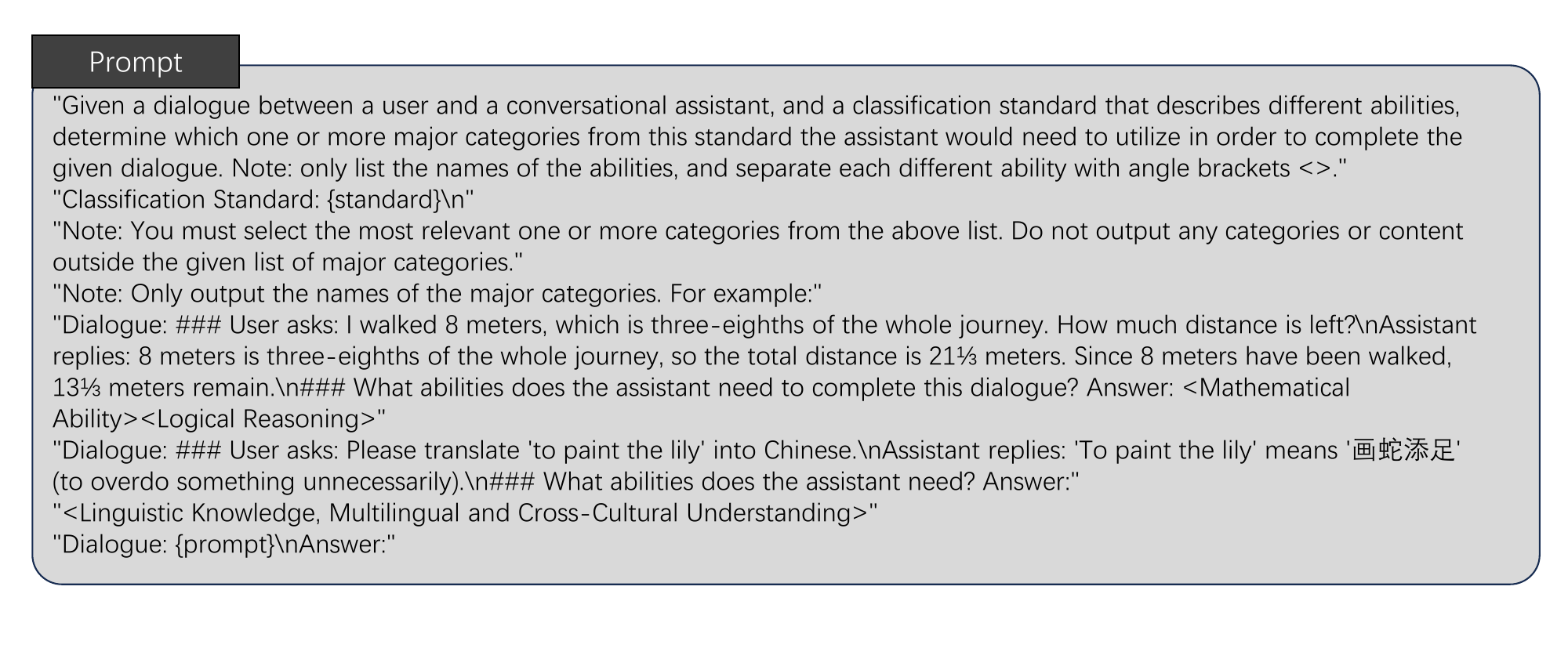}
    \caption{Prompt used for guiding the LLM to generate tags for given instruction.}
    \label{fig:prompt}
\end{figure*}

\paragraph{Fine-grained Tagging System}
Specifically, given an instance from the instruction dataset consisting of one or more {Query–Response} pairs, we concatenate them into a single string. We then use a carefully designed prompt to aviod generating over-detailed or over-coarse tags, as illustrated in Figure~\ref{fig:prompt}, to instruct LLMs to generate tags that describe the knowledge and skills required to complete the dialogue\footnote{We use Qwen-2.5-72B-Instruct as the tagging model to ensure high-quality annotations.}.
 

\paragraph{Tag Normalization}
Since LLMs may describe the same knowledge or skill using different expressions (e.g., ``math calculation'' vs. ``mathematical calculation''), we normalize tags based on semantic similarity. Specifically, we obtain embeddings for all tags using BGE \citep{chen2024bge}, and identify semantically similar tags whose cosine similarity exceeds an empirical threshold of $\lambda = 0.91$. Within each group of similar tags, we retain the one with the highest frequency as the representative.
To further refine the tag set, we apply DBSCAN \citep{ester1996density} clustering with parameters $eps =0.47$ and $min\_samples =2$ to merge closely related tags. These thresholds are chosen based on empirical observations. After normalization and clustering, tags with frequency less than 100 are considered noisy and removed, following prior work \citep{lu2023instag}. As a result, a total of 21,378 fine-grained tags are retained.





\begin{figure*}[h]
    \centering
    \includegraphics[width=0.9\linewidth]{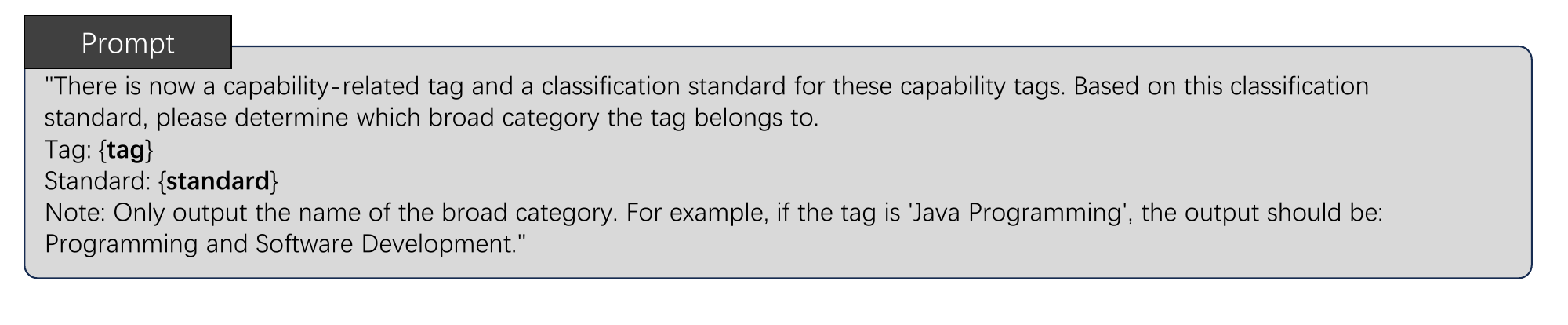}
    \caption{Prompt for categorizing a fine-grained tag into a domain-tag.}
    \label{fig:categorize_prompt}
\end{figure*}

\paragraph{Domain Tagging System}
Given the set of fine-grained tags, we select the top 1,000 most frequent ones, then use GPT-4 to automatically summarize them into a set of domain-level categories, so that we employ each induced category as a domain-level tag.
To establish a mapping between fine-grained tags and domain-level tags, we use the prompt shown in Figure~\ref{fig:categorize_prompt} along with the categorization criteria illustrated in Figure~\ref{fig:categorize_standad} of the Appendix. Based on this setup, we employ Qwen2.5-72B-Instruct \citep{qwen2.5} to generate the mappings between fine-grained and domain tags.


\begin{figure*}[h]
    \centering
    \includegraphics[width=0.9\linewidth]{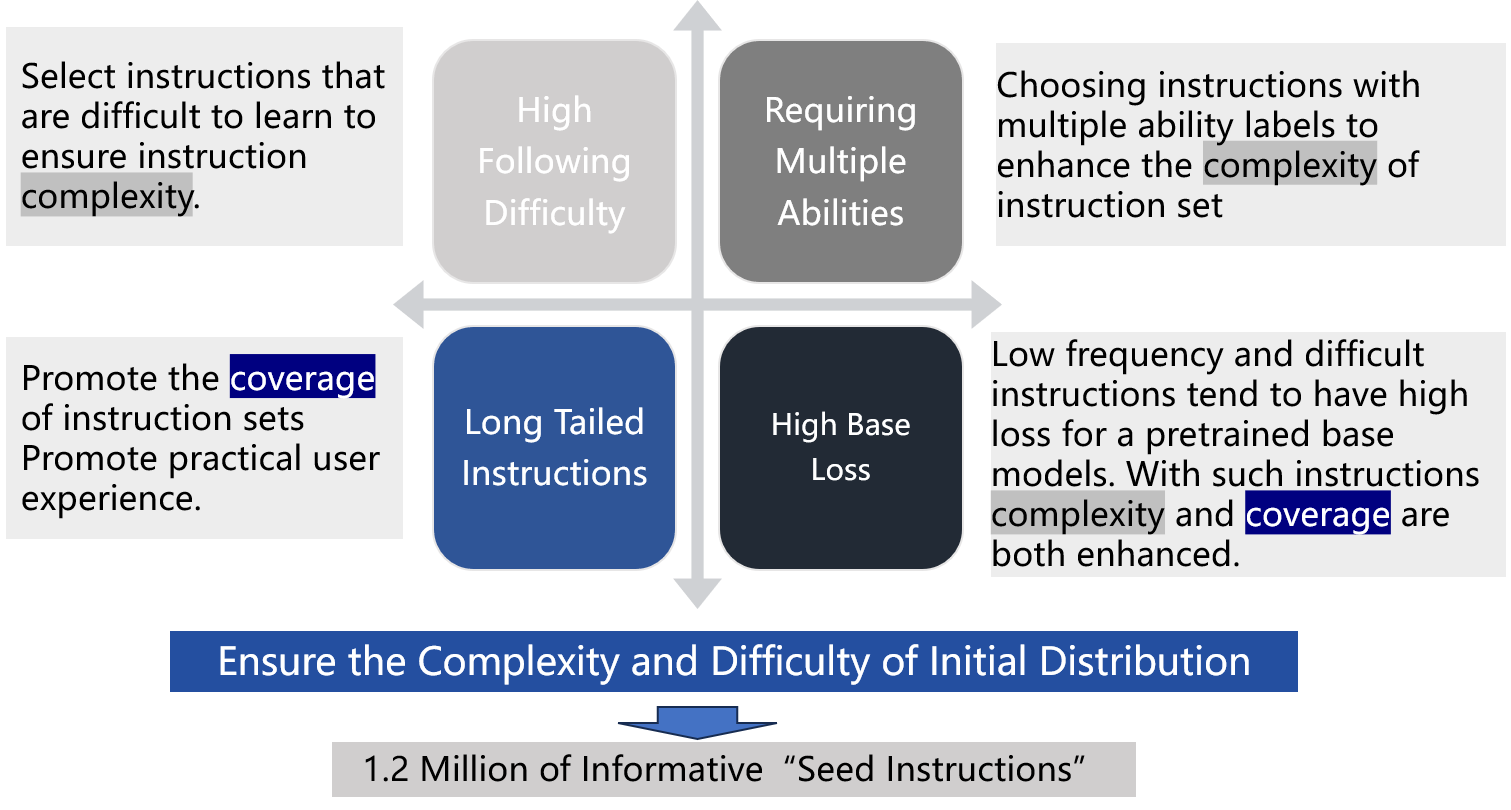}
    \caption{Standards for selecting informative seed instructions to enhance the depth and coverage of seed set.}
    \label{fig:data_selection}
\end{figure*}



\subsection{Informative Seed Instructions Selection}

The initial data distribution largely determines the effectiveness of subsequent data synthesis, thus selecting a set of informative seed instructions is crucial for synthesizing high-quality instruction data.
Given the abundance of existing instruction datasets, aimlessly synthesizing existing data is of limited value. 
In contrast, it is more meaningful to focus on instructions that are currently underrepresented in existing datasets (expanding coverage), or those that are relatively difficult and expose model limitations in reasoning or knowledge (enhancing depth)\citep{li2024quantity,shen2025long,huang2025musc,sun2024conifer}.
As shown in Figure~\ref{fig:data_selection}, we design four criteria to select informative instructions:







\textbf{1. Hard-to-Follow Instructions}:
Inspired by \citep{li2023quantity,qin2024infobench}, we select the 50k instructions with the smallest reduction in token-level average cross-entropy loss after fine-tuning, ensuring the retained instructions are challenging to follow.

\textbf{2. Long-Tail Instructions}:
To guarantee coverage, we include instructions that contain at least one fine-grained tag with a frequency below 200 in the seed set. Additionally, we randomly sample 30\% of the instructions that contain tags with frequencies in the range [200, 500].

\textbf{3. Multi-Skill Required Complex Instructions}:
Instructions associated with more than four fine-grained tags are included, as such instructions require more comprehensive reasoning or knowledge.

\textbf{4. Undertrained Instructions}:

These are instructions for which the base model performs poorly with a high loss value. This may either due to either a lack of exposure or inherent difficulty. We include 200k instructions with loss greater than $\text{mean(loss)} + 1.96 \times \text{std(loss)}$.

In particular, we use Llama-2-7B-base \citep{touvron2023llama} to compute the loss values for instructions.
To estimate the reduction ratio in token-level average cross-entropy loss, we fine-tune Llama-2-7B-base on a balanced instruction subset with 1,000 instructions per domain category.
Table~\ref{tab:dataset_summary} presents the sources of the seed data and the number of instructions retained after filtering.
By applying the above criteria sequentially, we select approximately 1.2 million informative instructions from the overall data pool.






\subsection{Evolutionary Algorithm for In-Distribution Generalization}

Based on the previously selected informative instructions as seeds, we employ an evolutionary algorithm to synthesize new instructions, so as to ensure the performance on instruction sets similar to the seed instruction distribution, meanwhile enlarge the coverage and increasing the depth of seed instructions. Specifically, we design a three-step evolution process:

\begin{figure}
    \centering
    \includegraphics[width=0.8\linewidth]{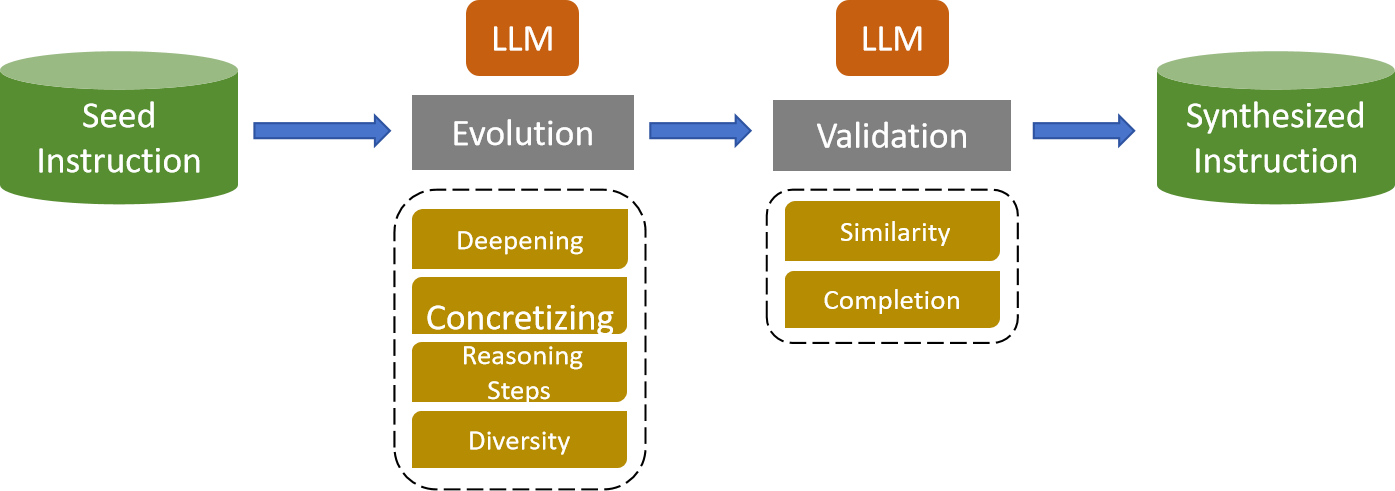}
    \caption{Process of the evolution-based data synthesizing.}
    \label{fig:evolution_process}
\end{figure}
1. Metadata-Guided Evolution:
In this step, we adopt the widely-adopted Evol-Instruct \citep{xu2023wizardlm} method. By randomly choose one of four dimensions—diversity, i.e., more reasoning steps, concretizing, deepening, and more diversified, to guide each instruction’s evolution using State-of-the-Art advanced large model as synthesizer.

2. Validation and Filtering:
Following \citep{tornberg2023chatgpt}, evolved instructions are evaluated by an advanced large model to judge the quality, as well as the similarity between the original seed instruction. 

3. Multi-Turn Dialogue Generation:
Following \citep{ding2023enhancing}, each valid instruction is used to generate 1–4 rounds of dialogue, with the AI assistant simulating different roles. 


\subsection{Deficiency Diagnosis and Defect-Driven Instruction Synthesis}

\begin{figure*}[h]
    \centering
    \includegraphics[width=0.7\linewidth]{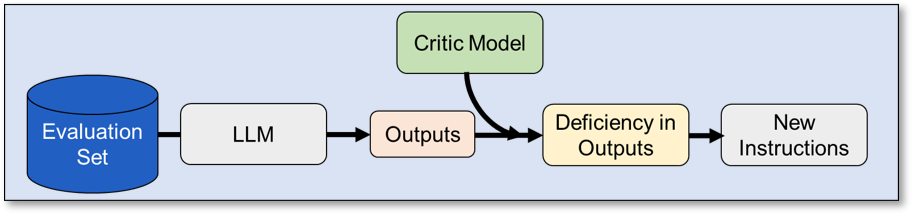}
    \caption{Illustration of the deficiency diagnosis process.}
    \label{fig:deficiency_diagnosis}
\end{figure*}



Despite comprehensive data collection efforts, many topics remain underrepresented, and the model may still fail to capture key knowledge and skills within the instruction set, leading to capability and knowledge deficiency.
While expanding the seed set or synthesizing more data via evolutionary methods can help, these approaches often produce redundant and inefficient results.
To address this issue, we adopt a more targeted strategy: directly diagnosing model deficiencies and synthesizing instructions specifically to fill those gaps.

\begin{figure}
    \centering
    \includegraphics[width=0.9\linewidth]{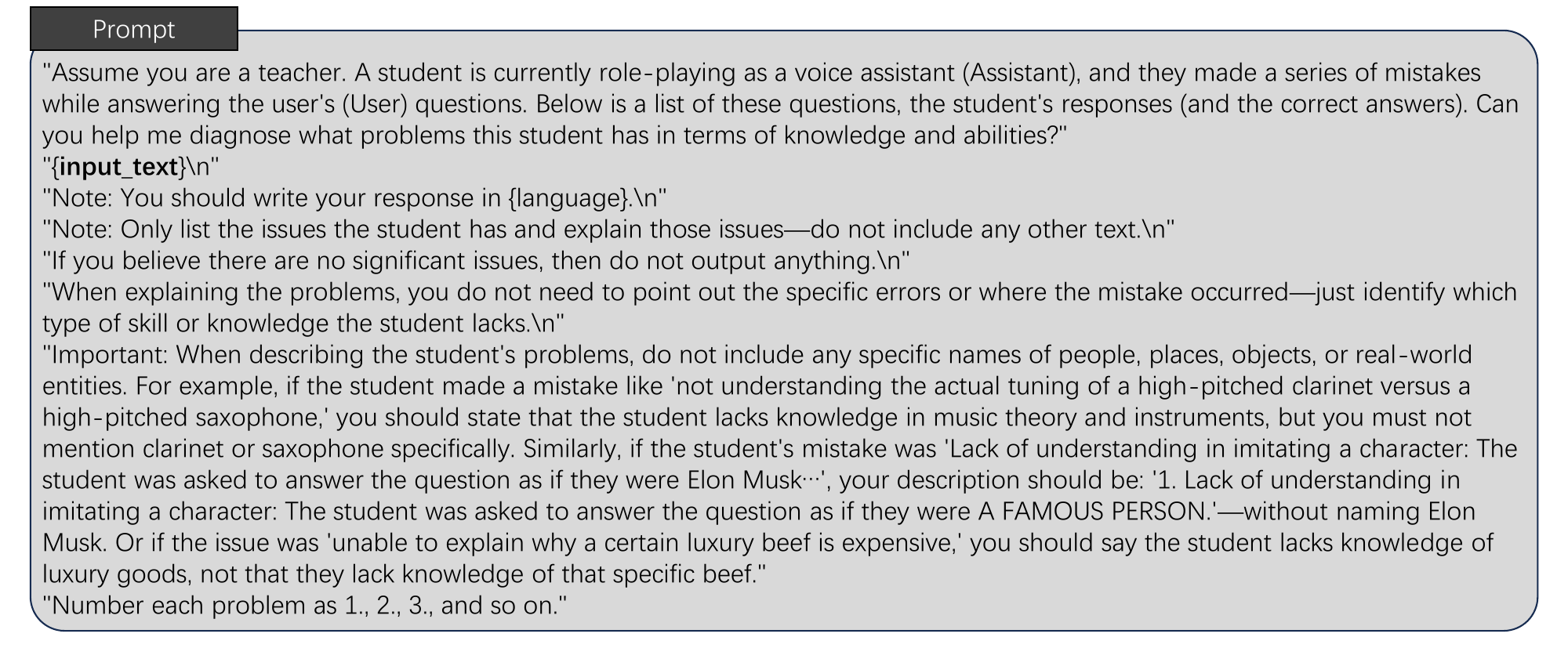}
    \label{fig:deficiency_prompt}
    \caption{Prompt for driving the oracle model for diagnosing deficiencies of finetuned model.}
\end{figure}

\begin{figure}
    \centering
    \includegraphics[width=0.9\linewidth]{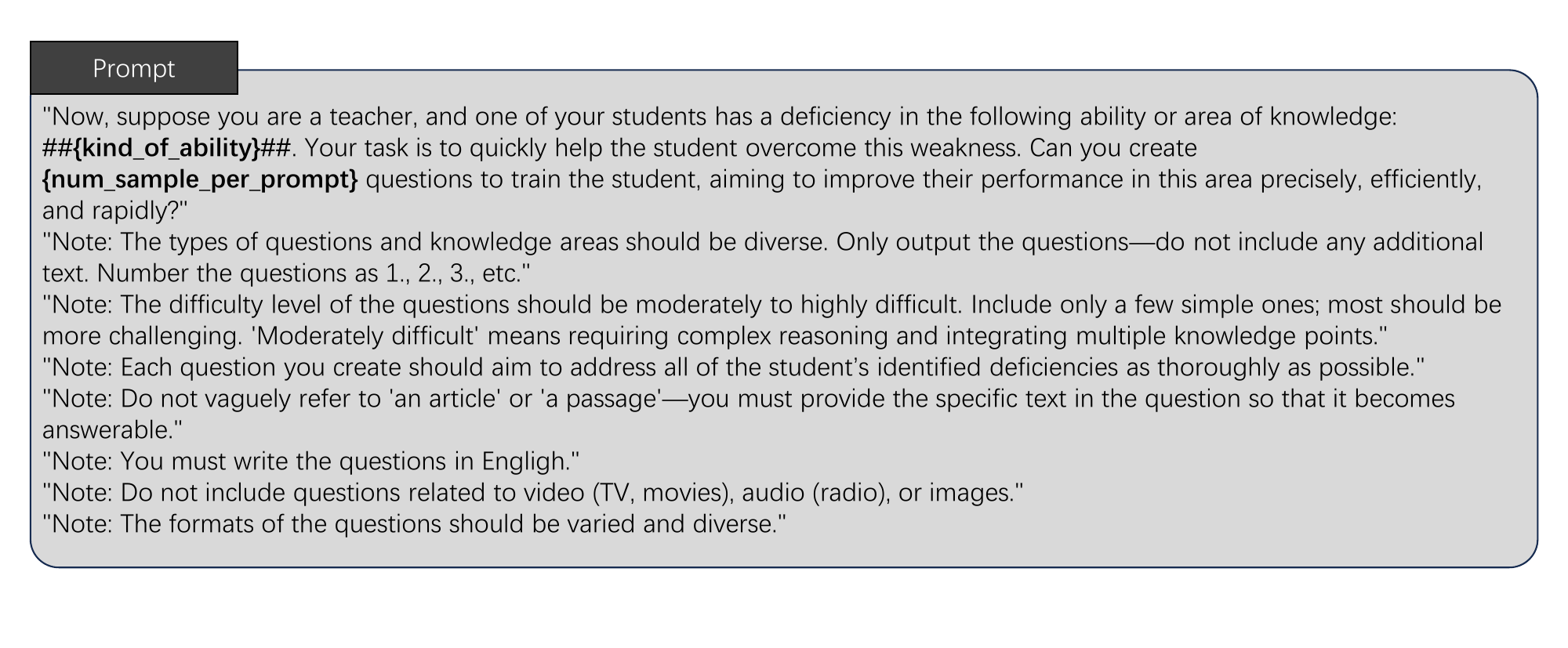}
    \label{fig:deficiency_based_syn_prompt}
    \caption{Prompt for driving the oracle model for target generating instructions to remedy deficiencies of the finetuned model.}
\end{figure}

As illustrated in Figure~\ref{fig:deficiency_diagnosis}, we first construct a diagnosis dataset $\mathcal{D} = {\mathcal{D}}_{i=1}^N$ drawn from the same distribution as the seed dataset, where each $\mathcal{D}_i = \{x_i, y_i\}$ represents an input query and its corresponding reference response. We then use the fine-tuned model $M_{\text{FT}}$ to generate a response $\hat{y}_i$ for each $x_i$.
Next, we compare $\hat{y}_i$ with $y_i$ using a State-of-the-Art advanced large Oracle model, identifying knowledge or deficiencies in $\hat{y}_i$ with the prompt shown in Figure~\ref{fig:deficiency_prompt}. Given the diagnosed issues, we then prompt the Oracle model to generate new instructions that specifically address these deficiencies using the prompt in Figure~\ref{fig:deficiency_based_syn_prompt}, to target generate instructions that remedy deficiencies within the model.

\subsection{Data Leakage Prevention}

Overlap between training and evaluation data can cause models to achieve inflated scores on familiar samples, leading to overestimated performance. We observe that many public instruction datasets include questions from common benchmarks like MT-Bench \citep{zheng2023judging} and AlpacaEval \citep{li2023alpacaeval}, which risks misleading evaluation.
To address this issue, we use the BGE model to compute semantic similarity between evaluation queries and instructions in the training data. This process involves multi-dimensional comparisons, including semantic understanding and structural features, to ensure accurate and comprehensive matching. Based on the similarity scores, we filter out overlapping entries, effectively preventing data leakage and preserving the reliability of model evaluation.

After the above procedures, we obtain 146,9391 new instruction samples.



\section{Performance Analysis}

\subsection{Experimental Setup}

We finetune open-source pretrained models Qwen2-7B-base and LLaMA3-8B-base on the InfInstruct-Sub dataset, and compare their performance against their respective official instruction-tuned and alignment-tuned versions. Evaluations are conducted on widely adopted LLM-judging-based benchmarks AlpacaEval 2.0 \citep{li2023alpacaeval}  and Arena-Hard-V0.1 \citep{li2024live}.
The reference model for AlpacaEval 2.0 is GPT-4-1106; on Arena-Hard-V0.1 is GPT-4-0314. 

\subsection{Experimental Results}

\begin{table}[h!]
\centering
\small
\caption{Performance comparison of various models across different benchmarks.}
\begin{tabular}{l|c|c|c}
\toprule
\textbf{Model} & \textbf{Datasize} & \textbf{AlpacaEval2.0} & \textbf{Arena-hard} \\ \midrule
GPT-4-1106 & -- & 50.0 & -- \\ \midrule
GPT-4-0314 & -- & 35.3 & 50.0 \\ \midrule
GPT-4-0613 & -- & 30.2 & 37.9 \\ \midrule
\midrule
Llama-3-8B  & & & \\
+Self-Instruct \citep{wang2023self}  & 100k & 7.21 & 4.0 \\
+ShareGPT \citep{zheng2023judging}              & 112k & 9.73 & 6.5 \\
+Evol Instruct \citep{xu2023wizardlm}         & 143k & 8.52 & 5.1 \\
+OpenHermes 1 \citep{OpenHermes1}          & 243K & 9.94 & 4.4 \\
+Tulu V2 Mix \citep{ivison2023camels}           & 326K & 9.91 & 5.4 \\
+WildChat \citep{zhaowildchat}              & 652k & 14.62 & 8.7 \\
+OpenHermes 2.5  \citep{OpenHermes2}       & 1M & 12.89 & 8.2 \\
+GenQA \citep{genqa_dataset}                  & 6.47M	 & 9.05 & 3.0 \\
+UltraChat \citep{ding2023enhancing}           & 208k & 8.29 & 3.6 \\
+Magipie \citep{xu2024magpie}            & 300k & 22.66 & 14.9 \\
\midrule
Llama-3-8B-Instruct & $>$10M &  22.92 & 20.6 \\
\midrule
Llama-3-8B+InfInstruct-Sub & 1.46M & \textbf{36.22} & \textbf{35.3}  \\
\midrule
\midrule
Qwen-2-7B-Instruct & - &  20.92 & 19.6 \\
\midrule
Qwen-2-7B+InfInstruct-Sub & 1.46M & \textbf{28.13} & \textbf{27.7} \\
\bottomrule
\end{tabular}
\label{tab:model_comparison}
\end{table}

Table~\ref{tab:model_comparison} shows the performance of the base model Llama3-8B and Qwen-2-7B fine-tuned with InfInstruct-Sub on AlpacaEval 2.0 and Arena-Hard-V0.1. 
Compared to earlier and concurrent instruction datasets, models fine-tuned with InfInstruct-Sub show improved performance, especially on ArenaHard, which focuses on more complex tasks. Moreover, comparison with performance using instruction collection OpenHermes 2.5 and GenQA, which have similar or larger sizes, shows that simply enlarging the size of the instruction set would not necessarily lead to performance improvement, particularly for harder instructions. This highlights the necessity of enhancing the depth of the instruction set. Moreover, on both Llama and Qwen, the performance outperforms official instruction-tuned counterparts. 
Notably, InfInstruct-Sub applies strict data leakage prevention and offers high coverage and difficulty, enabling improved generalization across diverse tasks. These results demonstrate the effectiveness of our proposed data construction framework to enhance model performance and its general applicability in enhancing instruction-following performance across different base models.

\section{Dataset Analysis}

\subsection{Dataset Distribution}

\begin{figure*}[h]
    \centering
    \includegraphics[width=0.8\linewidth]{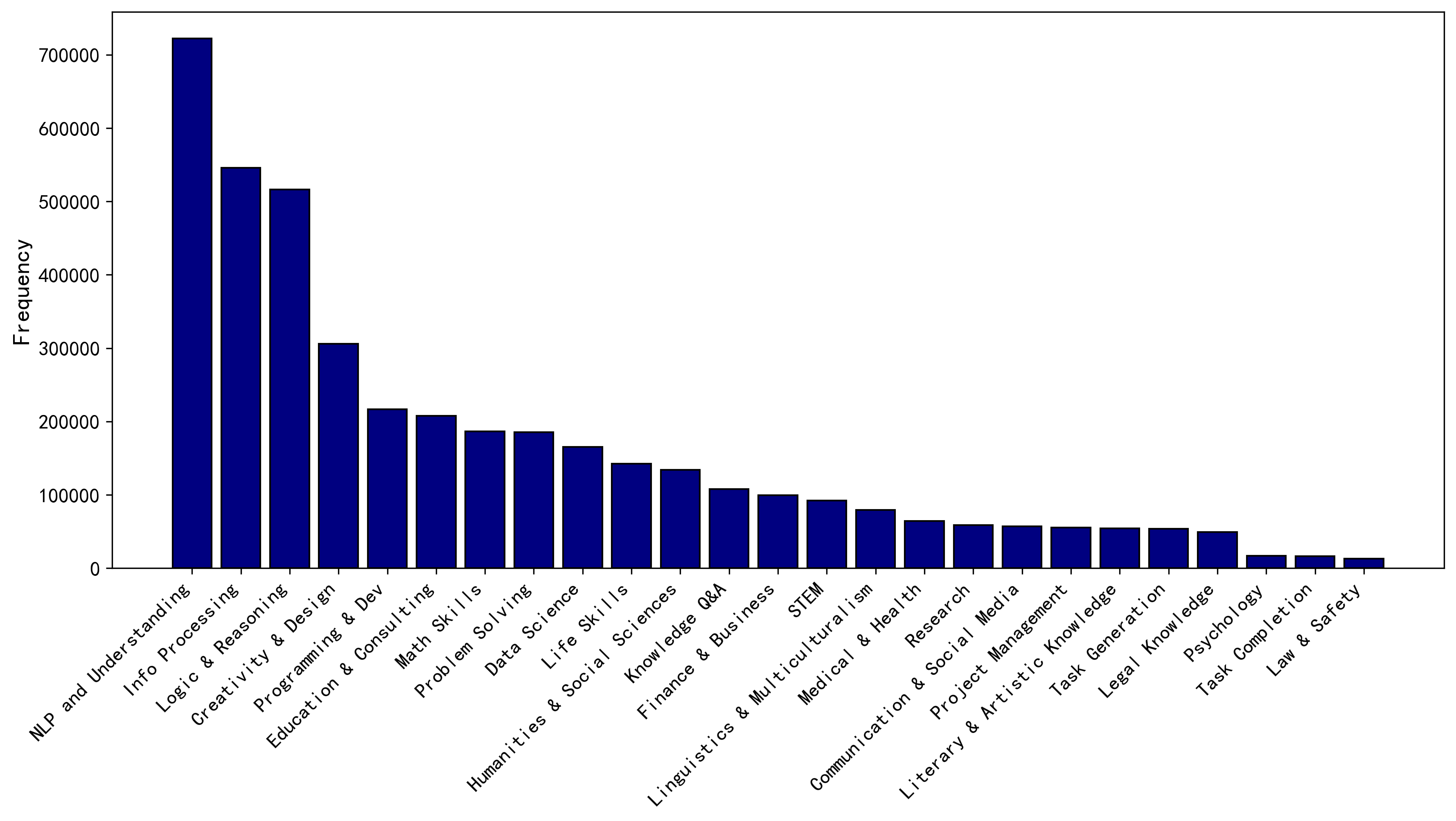}
    \caption{Illustration of the domain category distribution of the InfInstruct-Sub dataset.}
    \label{fig:cate_distribtion}
\end{figure*}

\begin{figure*}[h]
    \centering
    \includegraphics[width=0.9\linewidth]{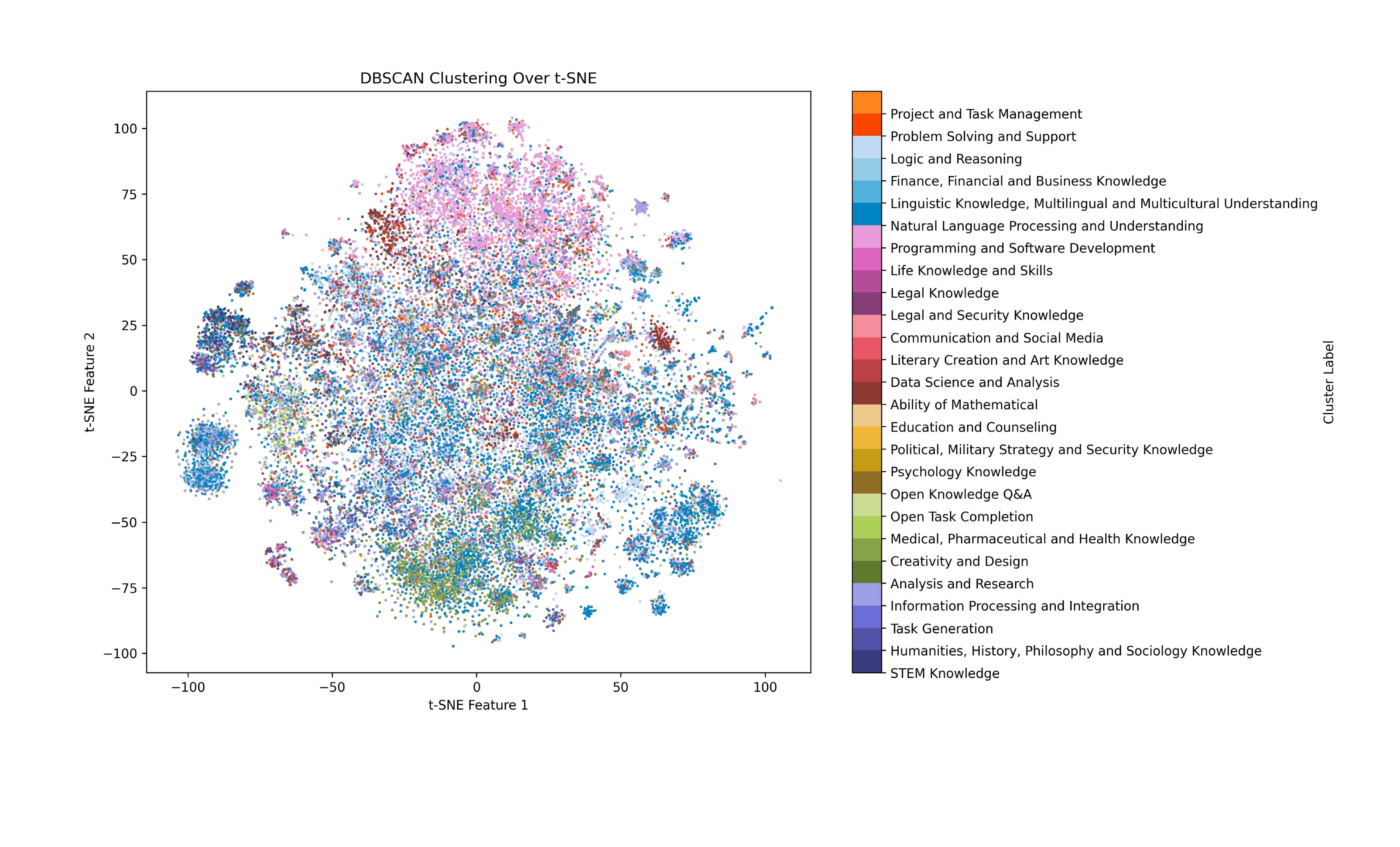}
    \caption{Illustration of the distribution within the semantic space of the synthesized instructions.}
    \label{fig:sem_distrib}
\end{figure*}


Figure~\ref{fig:cate_distribtion} shows the distribution of InfInstruct-Sub instructions across different domain labels.
Figure~\ref{fig:sem_distrib} visualizes the instruction data projected into semantic space using BGE as the text encoder, followed by t-SNE for dimensionality reduction. Overall, instructions with different domain labels are distributed in clearly separated regions, suggesting that instructions associated with different semantic tags carry distinct meanings. This supports the effectiveness of our tagging system.

\begin{figure*}[h]
    \centering
    \includegraphics[width=0.8\linewidth]{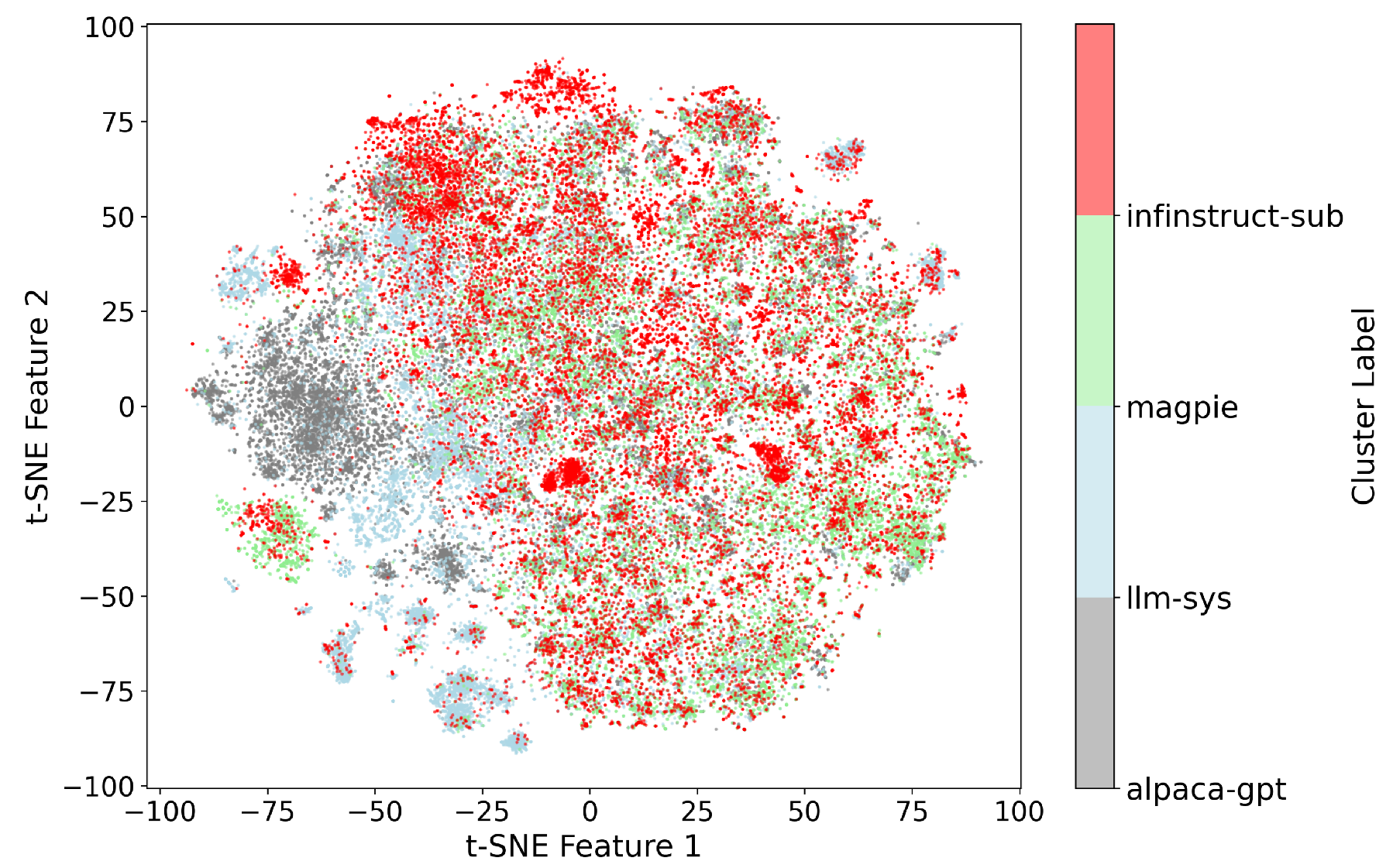}
    \caption{Distribution of the different instruction sets within the semantic space.}
    \label{fig:coverage_dataset}
\end{figure*}


We further compare the semantic coverage of InfInstruct-Sub with similar instruction datasets, including Alpaca\citep{taori2023stanford}, llm-sys\citep{zheng2023lmsys}, and Magpie \citep{xu2024magpie}. We sample 20,000 instructions from each dataset and apply the embedding and projection method same to abovementioned. As shown in Figure~\ref{fig:coverage_dataset}, InfInstruct-Sub exhibits a broader semantic spread, demonstrating that our strategy of selecting more low-frequency instructions as seeds effectively increases coverage in the synthesized data.


To further quantify the spatial distribution diversity of instruction sets, we use spatial entropy to evaluate the distribution uniformity of each dataset in semantic space. Spatial entropy is an extension of information entropy that quantifies the dispersion of points in space. Higher spatial entropy indicates a more uniform and diverse distribution.
Formally, we uniformly divide the 2D semantic space into $n$ discrete grids and compute the probability $p_i$ of a sample falling into the $i$-th grid. The spatial entropy is defined as:
\begin{equation}
H_{sp} = \sum p_i \log p_i
\end{equation}
In practice, $p_i$ is estimated as the proportion of samples falling into each grid cell. We divide the 2D space into a $200 \times 200$ grid and compute spatial entropy for each dataset accordingly. As shown in Table~\ref{tab:sp-dataset}, Infinity Instruct Subject achieves higher spatial entropy, indicating a more uniform and diverse distribution. This results from our targeted seed selection strategy, which emphasizes low-frequency and high-difficulty instructions, as well as instructions where the model underperforms, thus avoiding redundancy and enhancing the coverage and diversity of dataset.


\begin{table}
    \centering
    \caption{Spatial entropy of instructions within the 2D-semantic space.}
    \begin{tabular}{ccccc}
    \toprule
        Dataset & AlpacaGPT & llm-sys & Magipie & InfInstruct-Sub\\
    \midrule
        Spatial Entropy & 4.366 & 4.649 & 4.978 & \textbf{5.023} \\
    \bottomrule
    \end{tabular}
    \label{tab:sp-dataset}
\end{table}


\begin{figure*}[h]
    \centering
    \includegraphics[width=0.8\linewidth]{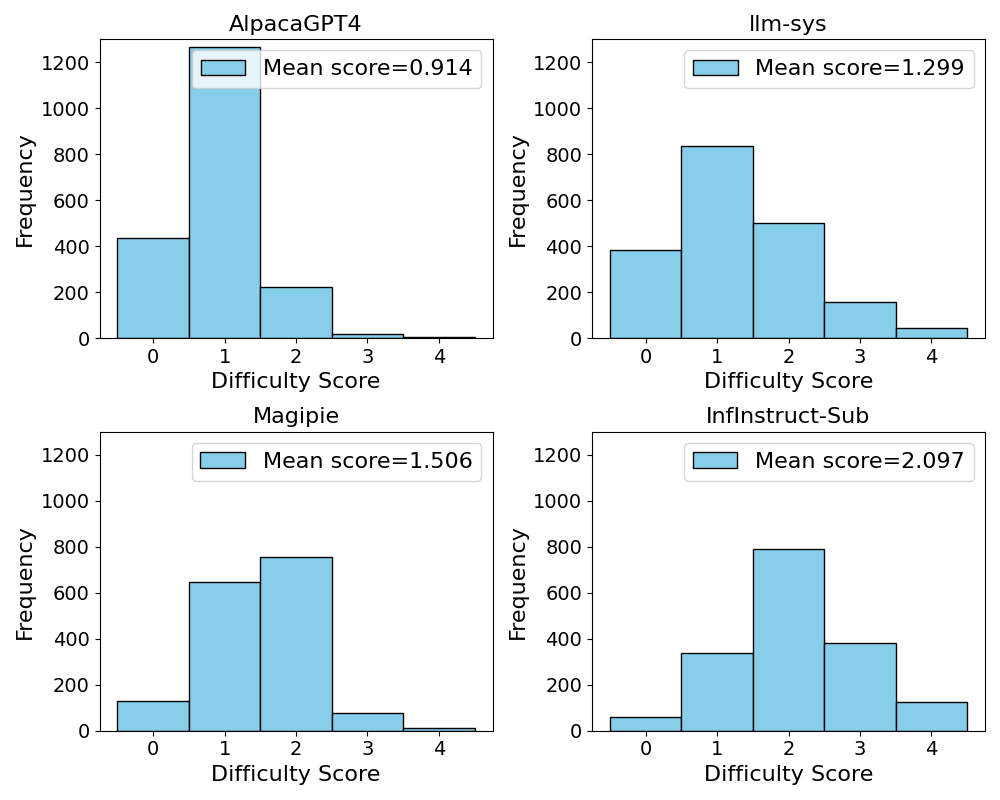}
    \caption{Distribution of difficulty scores of different instruction sets.}
    \label{fig:diff_distrib}
\end{figure*}


To evaluate the difficulty of synthesized data, we follow the approach proposed in Magpie \citep{xu2024magpie}, using a large language model to assign difficulty scores to instruction samples. The difficulty is rated on a five-point scale:
{'very easy': 0, 'easy': 1, 'medium': 2, 'hard': 3, 'very hard': 4}.
This evaluation process is independent of the method we use for selecting high-difficulty seed instructions.
Specifically, we employ Qwen-2.5-32B-Instruct as the scorer and randomly sample 1,500 instructions from each of the following datasets: Alpaca-GPT, LLM-Sys, Magpie, and InfInstruct-Sub.

Figure~\ref{fig:diff_distrib} shows the distribution of difficulty scores and the average score for each dataset. Compared to others, Infinity Instruct Subject contains more high-difficulty instructions and achieves a higher average difficulty score. This helps explain the superior performance of models fine-tuned InfInstruct-Sub on challenging benchmarks such as Arena-Hard (Table~\ref{tab:model_comparison}). It also validates the effectiveness of our seed selection strategy in enhancing dataset difficulty through targeted depth-oriented instruction synthesis.


\subsection{Performance Scaling with Depth and Coverage}


To further validate the effectiveness of our proposed method, we construct a series of instruction subsets from the synthesized data, each containing the same number of samples (20,000) but differing in depth and coverage. We define depth as the product of the logarithm of the instruction's label count and its token-level log loss of the base model. Coverage is measured by the logarithm of the number of non-empty grid cells occupied in the 2D semantic space described earlier.

\begin{figure}
    \centering
    \includegraphics[width=0.8\linewidth]{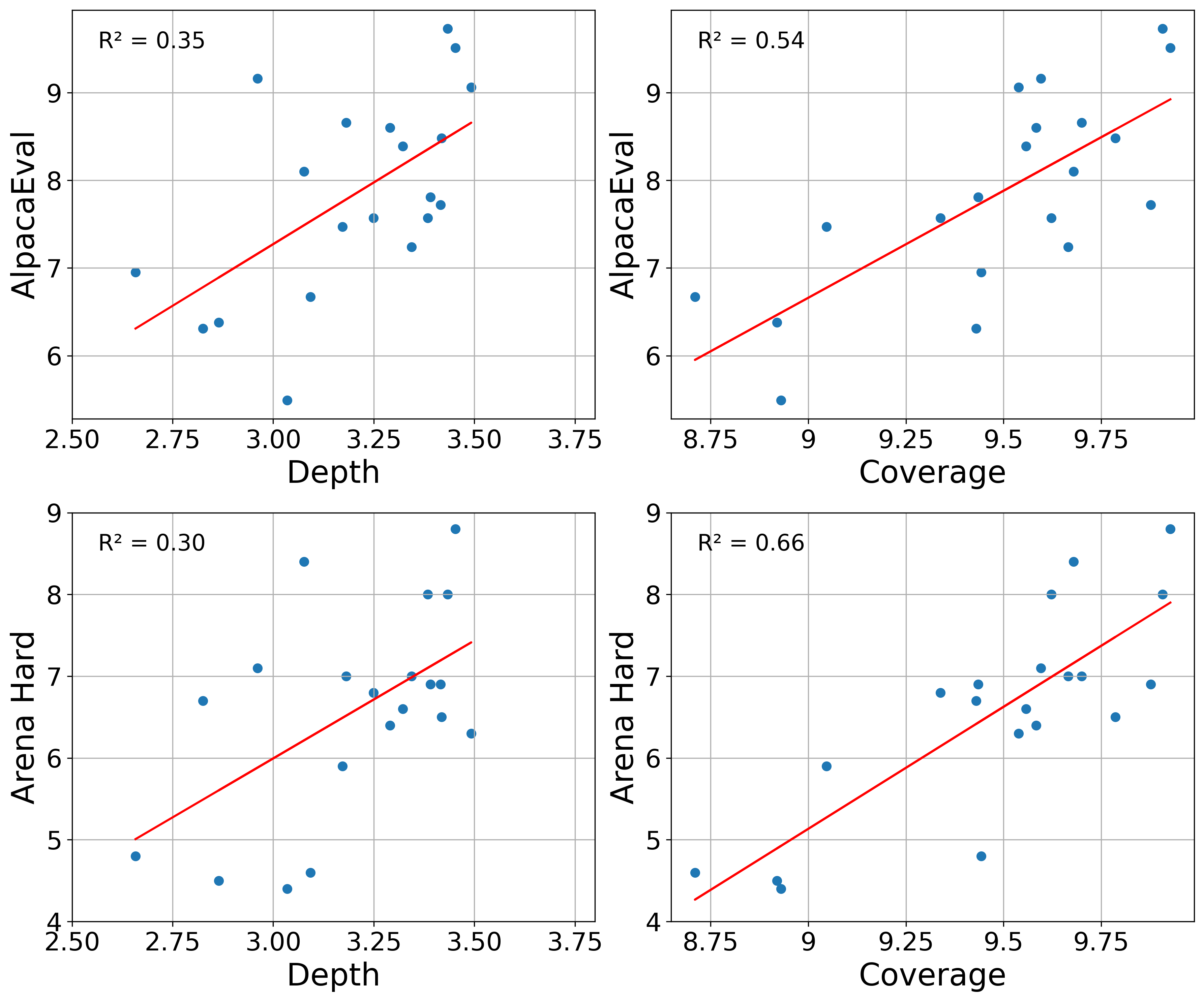}
    \caption{Performance of model finetuned with different coverage and depth.}
    \label{fig:scale_depth_coverage}
\end{figure}


Then, We fine-tune Llama3-8B on each subset and evaluate the aligned models on AlpacaEval and Arena-Hard. As shown in Figure~\ref{fig:scale_depth_coverage}, model performance increases consistently with instruction depth and coverage, even when dataset size is fixed. Intuitively, deeper instructions carry more informative signals, enabling the model to generalize better—including to simpler tasks. 
Previous studies have shown that instructions from different domains are difficult to substitute for each other. Therefore, it is necessary to enhance the coverage of instruction datasets to ensure model performance across various tasks. The correlation coefficient $R^2$ indicates that when the scale of the instruction dataset is around the tens of thousands, the model's performance may be more sensitive to the coverage of instruction coverage than to the depth of information.
These findings highlight the importance of expanding both depth and coverage of the instruction set, and further support the effectiveness of our strategy in achieving this goal during instruction synthesis, to continuously enhance the performance of models.


\subsection{Scaling Phenomenon in Label Connectivity Distribution}

\begin{figure*}[h]
    \centering
    \includegraphics[width=0.7\linewidth]{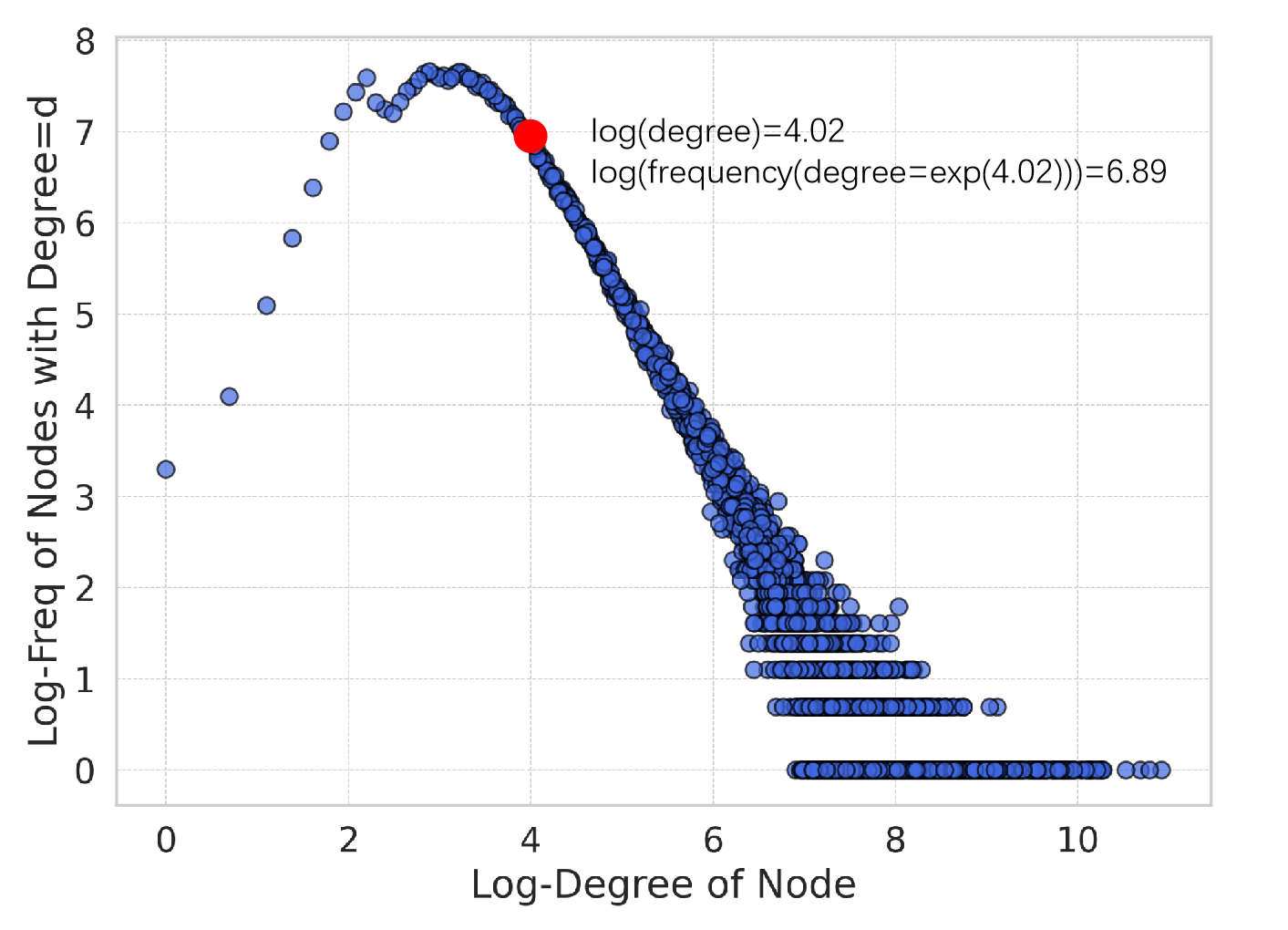}
    \caption{The connection degree of fine-grained tags with other tags exhibits a Power-Law distribution: Freq(degree=d)$~\sim \text{d}^{\gamma}$.}
    \label{fig:scaling}
\end{figure*}


During the data construction process, we observed an intriguing scaling law in the distribution of fine-grained tag connectivity. As shown in Figure~\ref{fig:scaling}, the frequency of tags with connectivity degree $d$ exhibits a clear negative log-log linear relationship:

\begin{equation}
   \text{log} [\text{Freq}(Degree=d)] \sim -\gamma \text{log} (d).
\end{equation}
or equivalently,
\begin{equation}
   P (Degree=d) \sim d^{-\gamma} .
\end{equation}

Here, the connectivity degree $d_i$ of tag $i$ is defined as:
\begin{equation}
    d_i=\sum_j \mathbb{I}_{ij}
\end{equation}
where $\mathbb{I}{ij} = 1$ if tag $i$ and $j$ co-occur in the same instruction. Intuitively, $d_i$ measures the degree of association between a given fine-grained tag (i.e., a type of knowledge or skill) and others in the dataset.


This pattern suggests that the underlying knowledge structure of instruction data may follow a scale-free topology similar to that found in complex networks such as the Internet. Prior studies have shown that in such networks, the probability of a node having degree $k$ follows:
\begin{equation}
P(k)\sim k^{-\gamma} 
\end{equation}
where $k$ denotes the degree of a node, and $P(k)$ represents the probability that a node with degree $k$ appears in the network.

In our case, this implies the existence of certain ``core'' knowledge or skills that frequently co-occur with a wide range of other rarer or more difficult skills. For instance, tasks involving [data analysis] often require both [mathematical reasoning] and [coding skills]. Similarly, [creative writing] tasks may simultaneously involve [commonsense knowledge] and [emotion analysis] to generate coherent and context-sensitive outputs.

The presence of this topology highlights the interdependent structure of knowledge in instruction data, which should be considered within the data construction, instruction selection, and model training process. This also highlights the necessity of understanding the distribution and interrelation of instruction contents \citep{zhao2025beyond}, and the necessity of a high-quality tagging system, which is the foundation of understanding such knowledge structure.

Moreover, prior work (e.g., \citep{tan2024information}) hypothesizes that the scaling law between model performance and dataset size is originated from the distribution of ``latent skills'' within the data, if such ``latent skills'' follow a power-law distribution. Our findings provide empirical support for this hypothesis, offering new insights into the origin of scaling laws between data distribution and model performance. We leave further exploration of this direction to future work, which further emphasizes the importance of the tagging system introduced in this paper.




\section{Related Work}




\subsection{Instruction Data Synthesis}

Instruction tuning is a fundamental approach to adapting base models to downstream tasks, and its effectiveness largely depends on the distribution of the instruction dataset.

Manually constructed datasets rely on experts to write instructions and responses, such as LIMA \citep{lima2023} and Dolly \citep{ding2023enhancing}. While these datasets are high in quality, they are expensive to scale due to the high cost of human annotation.


Semi-automatic approaches, like Self-Instruct\citep{wang2023self}, Alpaca\citep{taori2023stanford}, and Evol-Instruct\citep{muennighoff2023octopack}, expand a small set of human-labeled data using prompt engineering. While they improve scalability, the reliance on handcrafted prompts limits diversity and complexity.
Fully automatic methods (e.g., WebInstruct\citep{yue2024mammoth2}, backtranslation) extract instruction-like data from web documents with minimal human input, but often lack precise control over coverage and difficulty. Recent studies explore improvements from three angles:




\begin{itemize}
    \item Seed Selection and High-Information Filtering:
The choice of seed data largely determines the downstream diversity and quality. Traditional methods often use a few high-frequency, simple instructions, leading to limited coverage\citep{wang2023self}. Recent work addresses this by leveraging hierarchical tag systems and multi-dimensional metrics to select high-information seeds—i.e., rare, diverse, and complex instructions that better represent undercovered regions of the task space\citep{tu2024resofilter,he2025fine,xie2023data}. This ensures that synthesized data can expand both the coverage (domain/task diversity) and depth (instruction complexity) of the dataset.

    \item Evolution-Based Instruction Generation:
To address the need for more challenging instructions, evolution-based methods iteratively expand seed data using strategies like genetic algorithms \citep{xu2023wizardlm} or model feedback loops\citep{gu2025llm}. These approaches increase instruction complexity and reasoning depth, while dynamically adjusting generation based on model performance. They also help avoid redundancy by diversifying synthesis paths.

    \item Instruction Synthesis Strategies:
Most existing synthesis methods rely on prompting large language models with handcrafted or auto-generated templates\citep{wang2023self, xu2023wizardlm}. While effective, this can lead to templated and repetitive outputs. Magpie\citep{xu2024magpie} proposes prompt-free instruction synthesis via autoregressive alignment, which improves fluency and semantic richness. Similarly, Web Reconstruction \citep{jiang2025instruction} leverages real-world text to generate instruction-style data with higher realism and contextual diversity.

\end{itemize}



Despite these advances, many datasets still underrepresent complex, multi-step, or domain-specific instructions. Few existing works explicitly address the coverage and depth of instruction distribution in a unified manner. This limits model generalization in scenarios requiring deep reasoning, long-range context, or adaptation to unfamiliar domains \citep{yu2024survey, zhang2025out}.
Moreover, while some studies explore feedback-driven data augmentation or generative instruction synthesis \citep{le2022coderl}, most lack a closed-loop framework that jointly optimizes coverage, difficulty, and model adaptation. The interplay between seed selection, model feedback, and evolution remains underexplored.

To address these gaps, we propose a unified instruction data construction framework that systematically expands both coverage and complexity. By integrating hierarchical tagging, high-information seed filtering, evolution-based synthesis, and deficiency-driven augmentation, our approach enables structured, scalable, and adaptive instruction data generation, laying the foundation for robust and generalizable instruction-tuned models.



\subsection{Model Self-Improvement}

Model self-improvement aims to iteratively enhance model capabilities through self-generated data or feedback signals. Representative approaches include self-refinement \citep{madaan2023self, zhou2024cycle}, multi-round generation and evaluation loops, and performance-driven data augmentation \citep{yang2025knowing}. These methods enable models to identify and address their own weaknesses by generating more challenging training samples, thereby facilitating continual improvement.


To ensure data quality, recent studies introduce deficiency diagnosis mechanisms \citep{lighterness2024data}, which analyze model performance on downstream tasks to detect knowledge gaps or skill deficiencies. These insights guide the targeted synthesis of training data\citep{miller2025comparison}. By incorporating such feedback into the training loop, models and datasets can co-evolve—improving both model performance and data coverage over time\citep{bauer2024feedback}.


Beyond simply increasing instruction quantity, recent research emphasizes instruction coverage (task/domain diversity) and depth (reasoning complexity). Several works have explored combining hierarchical tagging with complexity control to synthesize increasingly challenging and diverse instructions via evolutionary algorithms \citep{lu2023instag,zhao2025beyond}. These strategies enhance the cognitive depth of training data and promote models’ abilities in complex reasoning and compositional generalization.

\section{Conclusion}

In this paper, we propose a novel framework, for constructing high-quality instruction datasets that continuously expand both the coverage and complexity of instruction data. By combining a hierarchical multilingual tagging system, informative seed instruction selection, evolutionary data synthesis, and model deficiency diagnosis, our framework enables generating instructions that effectively address gaps in coverage and complexity within existing instruction datasets. We demonstrate the utility of this framework through the construction of the Infinity Instruct Subject dataset, which contains over 1.5 million high-quality instructions, and show that it improves the instruction-following capabilities of foundation models across benchmarks. Further analyses show that our methodology can obtain instructions with enhanced coverage and depth. Moreover, the data distribution analysis uncovers interesting scaling laws in instruction data, suggesting a scale-free relationship between the co-occurrence of instruction tags and their frequency. These findings open up new avenues for understanding the internal knowledge structure of instruction datasets and offer insights into the scaling behavior of large models, which can be leveraged to improve training efficiency and model performance.

\bibliographystyle{unsrtnat}
\bibliography{Ref}

\appendix
\section{Appendix}
\setcounter{table}{0}   
\setcounter{figure}{0}

\begin{table}[h!]
\centering
\small
\begin{tabular}{l|r}
\toprule
\textbf{Raw Dataset} & \textbf{Numbers of Rows} \\ \midrule
Alpaca GPT4 data & 13,490 \\ \midrule
Alpaca GPT4 data zh & 32,589 \\ \midrule
Baize & 14,906 \\ \midrule
BELLE Generated Chat & 43,775 \\ \midrule
BELLE Multiturn Chat & 210,685 \\ \midrule
BELLE 3.5M CN & 312,598 \\ \midrule
databricks-dolly-15K & 10,307 \\ \midrule
LIMA-sft & 712 \\ \midrule
CodeContest & 523 \\ \midrule
LongForm & 3,290 \\ \midrule
ShareGPT-Chinese-English-90k & 8,919 \\ \midrule
UltraChat & 237,199 \\ \midrule
Wizard evol instruct zh & 44,738 \\ \midrule
Wizard evol instruct 196K & 88,681 \\ \midrule
BELLE School Math & 38,329 \\ \midrule
Code Alpaca 20K & 13,296 \\ \midrule
WildChat & 61,873 \\ \midrule
COIG-CQIA & 45,793 \\ \midrule
BAGEL & 55,193 \\ \midrule
DEITA & 10,000 \\ \midrule
Summary & 1,342,427 \\ \bottomrule
\end{tabular}
\caption{Dataset Summary}
\label{tab:dataset_summary}
\end{table}

\begin{figure*}
    \centering
    \includegraphics[width=0.9\linewidth]{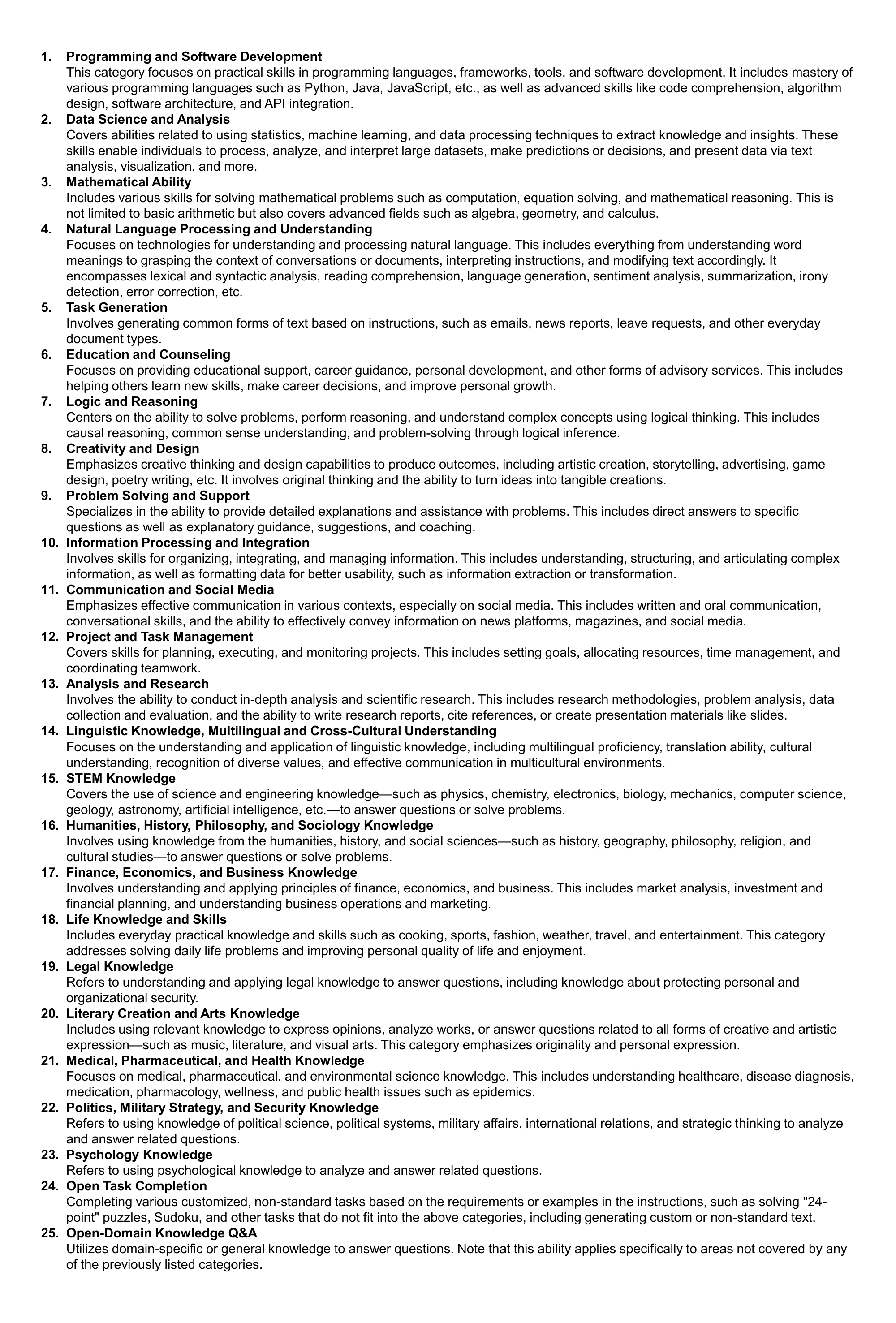}
    \caption{Classification standard for categorizing a fine-grained tag into a domain-tag.}
    \label{fig:categorize_standad}
\end{figure*}
\end{CJK}

Given a dialogue between a user and a conversational assistant, and a classification standard that describes different abilities, determine which one or more major categories from this standard the assistant would need to utilize in order to complete the given dialogue. Note: only list the names of the abilities, and separate each different ability with angle brackets $<>$.
Classification Standard: {standard}

\end{document}